# A novel approach to sentiment analysis in Persian using discourse and external semantic information

*Rahim Dehkharghani, Faculty of Engineering, University of Bonab, Bonab, Iran
rdehkharghani@bonabu.ac.ir
Hojjat Emami, Faculty of Engineering, University of Bonab, Bonab, Iran
emami@bonabu.ac.ir

**Abstract**
Sentiment analysis attempts to identify, extract and quantify affective states and subjective information from various types of data such as text, audio, and video. Many approaches have been proposed to extract the sentiment of individuals from documents written in natural languages in recent years. The majority of these approaches have focused on English, while resource-lean languages such as Persian suffer from the lack of research work and language resources. Due to this gap in Persian, the current work is accomplished to introduce new methods for sentiment analysis which have been applied on Persian. The proposed approach in this paper is two-fold: The first one is based on classifier combination, and the second one is based on deep neural networks which benefits from word embedding vectors. Both approaches takes advantage of local discourse information and external knowledge bases, and also cover several language issues such as negation and intensification, andaddresses different granularity levels, namely word, aspect, sentence, phrase and document-levels. To evaluate the performance of the proposed approach, a Persian dataset is collected from Persian hotel reviews referred as *hotel reviews*. The proposed approach has been compared to counterpart methods based on the benchmark dataset. The experimental results approve the effectiveness of the proposed approach when compared to related works.

## 1- Introduction

Sentiment analysis is the process of using text analysis, natural language processing, and computational linguistics to identify and extract polarities toward entities such as topics, events, individuals, issues, services, products, organizations, and their attributes [1]. Polarity usually refers to positivity and negativity; however, it could investigate other types of emotion such as fear, excitement, joy, etc. Data used in sentiment analysis systems are usually in textual format but other types of data such as video, image, or audio could be analyzed. This research area has attracted many researchers in recent decades.

Sentiment analysis is a domain and language-dependent task. Words or phrases may carry different polarities in different domains. For example, the word "big" has positive polarity for the "room size" in the hotel domain, while it has negative polarity for the "battery size" in the camera domain. There exist a big deal of research on sentiment analysis in some languages; however, Persian is one of the less-studied languages in this field [2]. Persian is a member of the Indo-European languages. Over 110 million people speak Persian in different countries such as Iran, Afghanistan, Tajikistan, and Uzbekistan, which constitute 1.5% of the world's population [2] who produce the bulk of Persian content on the web and social media. Figure 1 illustrates the growth of Persian digital content over the past years and the prediction of its growth for the future. As shown in Figure 1, the volume of the Persian content has increased at a steady rate over the past years. This is expected also for the future. While some methods in natural language processing can be shared among different languages or contexts, sentiment analysis systems need to be specialized for the language or domain of interest.

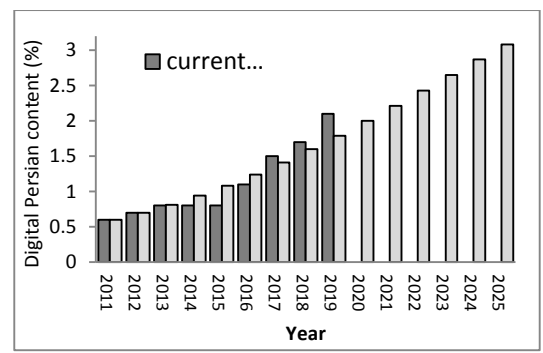

Figure 1. The growth of Persian content on the web and various social media[1]

---

[1] The data used for drawing the current status of Persian content on the Internet are driven from "Usage of content languages for Websites", [www.W3Techs.com], Retrieved 30 Nov. 2019

* Corresponding Author

As an element of research, in this paper, we investigate sentiment analysis in Persian and propose an approach that covers different granularity levels and language issues. In the second fold of proposed approach, we also use word embedding vectors in a deep neural network

Persian is a challenging language for sentiment analysis due to its special and different nature such as linguistic phenomena, shortage of appropriate natural language processing (NLP) tools and underlying linguistic resources. Some challenges include misspelling, word spacing, and use of informal words. Most Persian letters have multiple forms of writing and there are a lot of frequent exceptions in words order. The use of wide variety of declensional suffixes and some imported sounds from Arabic which may be written or ignored, result in various forms of writing. We addressed most of the above –mentioned issues by the help of existing tools for Persian such as Hazm, but some other issues such as dealing with informal writing form are ongoing problems in Persian and therefore not solved in this paper.

To summarize, the contributions of this paper are as follows.

- A classifier combination approach is proposed for sentiment analysis, which covers language issues such as negation, intensification, and also different granularity levels such as aspect and phrase levels.
- Resolving co-referent mentions in the pre-processing phase to improve the performance of sentiment analysis.
- A combination of three Persian polarity lexicons, namely, SentiFars, PerSent and LexiPers is used for feature extraction, which increases the document classification performance and robustness.
- Pre-trained word embedding vectors generated by FastText are used for polarity classification in a deep neural network.
- Comprehensive experiments are performed to evaluate the potential of the proposed approach. The experiments show the superiority of the proposed approach when compared with counterparts.

The remaining of this paper is organized as follows. Section 2 discusses the related works and their merits and drawbacks in the field of sentiment analysis. Section 3 explains the proposed method. Section 4 presents the experiments and compares the proposed method with a baseline and state of the art. Finally, Section 6 concludes the paper and suggests future directions.

## 2- Related work

Sentiment analysis has attracted a lot of research interest in recent years, especially in the context of web and social media. Sentiment analysis can be examined in different granularity levels: document-level [3], sentence-level [4], phrase level [5], aspect level [6], and word-level [7]. On the other hand, the sentiment analysis methods can be grouped into three main categories [1]: machine learning methods, lexicon-based methods, and hybrid methods.

Machine learning methods are more popular as they achieve encouraging results on sentiment analysis. An important branch of machine learning methods is a supervised classification, which can be automatically trained and applied on various domains. Supervised methods such as Naive Bayes (*NB*), Maximum Entropy (*ME*), Artificial Neural Networks (*ANNs*), and Support Vector Machines (*SVM*) have achieved great success in sentiment analysis. Liu et al. [8] proposed a supervised multi-class sentiment classification method based on SVM and an improved one-vs.-one strategy. After mapping the training texts to feature vectors, the information gain method is used to extract important features for multi-class sentiment grouping. To determine the sentiment class of a given piece of text in the test set, a confidence score matrix of multiple SVM classifiers is constructed. Then, the polarity of this text is identified using the one-vs-one strategy. Tang et al. [9] proposed Neural Network models including Conv-GRNN and LSTM-GRNN for document-level sentiment classification. The model first learns sentence representation with a convolution neural network. Then, the semantic of sentences and their relations are adaptively encoded in document representation with gated Recurrent Neural Network. Parlar et al. [10] introduce a new feature selection method, called query expansion ranking (QER) for sentiment analysis from review texts. QER is based on query expansion term weighting methods. The main drawback of the supervised methods is that they require a relatively large training set. To overcome this problem, unsupervised methods are devised.

In unsupervised methods, there is no need for training data to train the sentiment analysis system. Riz et al. [11] proposed a phrase-level sentiment analysis method to identify customer preferences by analyzing subjective reviews. The authors extracted the polarity of words to find out the intensity of each expression using *k*-means clustering algorithm. Suresh and Gladston [12] presented a novel fuzzy clustering method to analyze tweets regarding the sentiments of a particular brand. Feature extraction and feature selection are two bottlenecks of unsupervised techniques, as these tasks directly affect the classification performance.

In lexicon-based methods, the polarity of a sentence or document is estimated based on the polarity of its components (words/ phrases) using polarity lexicons. Polarity lexicons contain a (generally large) set of words/ phrases, which express individuals' feelings and opinions towards an issue by using quantitative values. For example, the polarity score of the word "happiness" in SenticNet, a polarity lexicon in English, is +0.14. The approach of lexicon-based techniques towards sentiment analysis is unsupervised, because they do not require prior training phase to classify data. Turney [13] employed a set of patterns of tags for extracting two-word phrases from reviews. Then PMI-information retrieval (PMI-IR) method is employed to determine the semantic orientation of review by issuing queries to a search engine. Agarwal et al. [14] performed polarity classification of Tweets. The authors employed five different combinations of features on unigrams, senti-features, and tree kernel. They evaluated the proposed method with 11,875 manually annotated tweets.

Some researchers combined lexicon-based and machine learning methods to extract sentiment from data sources. For example, Basari et al. [15] proposed a hybrid method that is composed of SVM and Particle Swarm Optimization (PSO) for sentiment analysis of movie reviews. PSO is used to find out the best parameters to solve the dual optimization problem. El Rahman et al. [16]proposed a hybrid approach that combines lexicon-based classification and unsupervised clustering methods to extract the polarity of real data collected from Twitter.

Most research interest has focused on the English language. Only few studies have been performed on sentiment analysis for resource-lean languages such as Persian[17]–[25]. Since English and Persian have different characteristics, in order to apply proposed approaches for English, on Persian, they need to be modified first. Therefore, most proposed approaches for Persian have been specialized for this language. Takhshid and Rahimi [20] proposed a rule-based method for detecting negative words in Persian. Vaziripour et al. [26] classified the sentiment of individual tweets to find out the opinions of their authors towards a number of trending political topics. They used an SVM classifier with Brown clustering for feature selection, which yielded an accuracy of 70%.

Dashtipour et al. [22] used two deep learning models including deep auto-encoders and deep convolutional neural networks (CNNs) to extract sentiments from Persian movie reviews. The input data is passed to pre-processing phase to perform tokenization, normalization and stemming on text. Then, the pre-processed text is concerted to word vectors using Fasttextlibrary. For classification of reviews, multilayer perceptron (MLP), auto-encoders and CNNs are used.

Basiri et al. [23]proposed a new sentiment aggregation method based on the cross-ratio operator. They investigated the effects of the sentiment lexicon, aggregation level, and aggregation method on the sentiment polarity and rating classification of Persian reviews. They conclude that the review-level aggregation can improve rating classification, but it does not have a positive impact on polarity classification.

Dashtipour et al. [24] proposed a hybrid framework for concept-level sentiment analysis in Persian, that integrates deep learning and linguistic rules to optimize polarity detection. When a pattern is triggered, the framework allows sentiments to flow from words to concepts based on symbolic dependency relations. When no pattern is triggered, the framework switches to its sub-symbolic counterpart and leverages deep neural networks (DNN) to perform the classification.

Roshanfekr et al. [25]investigated deep learning techniques for classifying documents based on their sentiment polarity. They compared deep leaning methods with support vector machine(SVM)-based methods. The results show the superiority of using deep learning methods when compared with its counterparts. The success of their method is due to using a word vector representation which solves most of the challenges arises from different writing styles in Persian, and the lack of the datasetby utilizing unsupervised methods.

Some research works focus on generating polarity lexicons [27] and corpora [28] for Persian. There exist three publicly available sentiment lexicons for Persian, which have been used in the current work [29], [30], and[27]. In [30], Dashtipour and colleagues explain the entire process of building a manually annotated sentiment corpus, named PerSent. This corpus includes about 1500 Persian entries (words and phrases), each of which has a polarity score between -1 and 1 and a Part of Speech (POS) tag. In [27], Sabeti and his colleagues propose a new graph-based method for seed selection and expansion to generate general-purpose polarity lexicons, which results in generating a Persian polarity lexicon named LexiPers including over 6000 entries. The authors benefit from dictionary-based and corpus-based approaches. In [29], as our previous work, a translation-based approach, based on classifier combination is proposed to generate a polarity lexicon for Persian, named SentiFars, which includes over 2600 entries with three polarity scores-positive, negative, objective, summing up to 1-for each entry. Other issues such as feature selection[18], [31] and feature extraction [21] for sentiment analysis have been also investigated for Persian.

# 3- Proposed approach

The proposed approach takes Persian reviews as input and classifies them into one of the three classes: positive, negative, and neutral. This approach is two-fold: classifier combination approach and deep learning approach. The former approach is composed of four main steps: pre-processing, feature extraction, feature integration, and grouping. The latter approach consists of pre-processing and classification steps. The preprocessing step is almost the same in both approaches but other steps are different; Figure 2 illustrates the state diagram of the proposed approach.

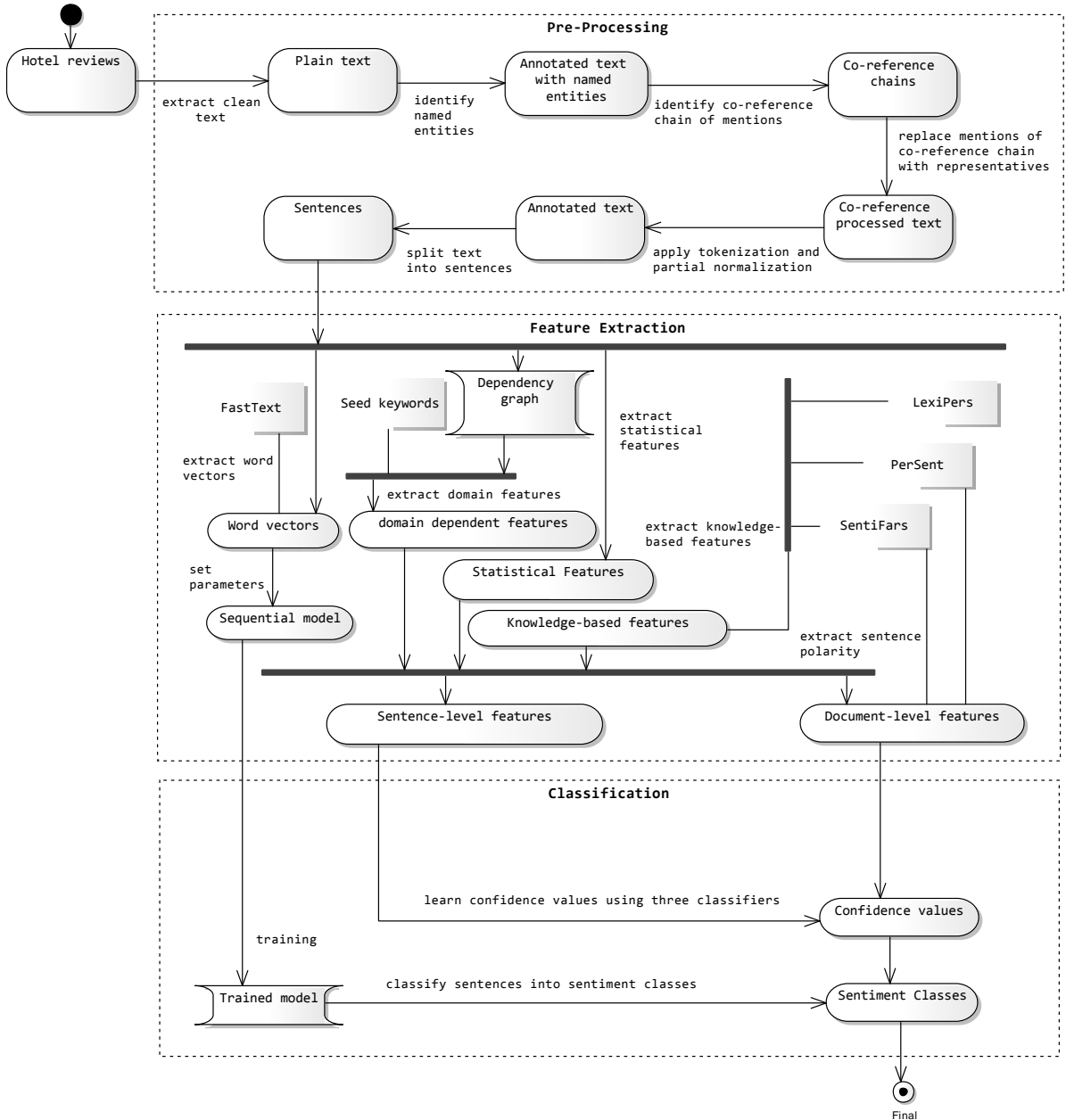

Figure2: the state diagram of our proposed approach

## 3-1- Pre-processing

Pre-processing takes as input, documents $D = \{D_1, D_2, ..., D_n\}$ and prepares them for the next processes. Pre-processing consists of four steps including co-reference resolution, sentence segmentation, tokenization, and partial normalization. The last step includes cleaning the words, lowercasing, and plural to singular transformation

### 3-1-1- Co-reference resolution

In this phase, pronouns and entity mentions are replaced by their corresponding representative mentions in the text. To resolve co-references, first, coarse-grained entity types including person, location and organization are annotated using a multi-lingual named entity recognizer [32]. Then, annotated documents are passed to a rule-based co-reference resolution module [33] to resolve co-referent mentions. This module identifies co-reference chains for all the entities mentioned in the document. The mentions within every co-reference chain are replaced with their referent mention. For example in the following sentences,

کیف چرمی خیلی زیبا است. آن را خیلی دوست دارم.
The leather case is very beautiful. I like it so much.
The pronoun "it/آن" is replaced with mention "کیف چرمی /the leather bag" which results in the following sentence:

کیف چرمی خیلی زیبا است. کیف چرمی را خیلی دوست دارم.
The leather bag is very beautiful. I love the leather bag.

### 3-1-2- Sentence segmentation

In this step, each document $D_i \in D$ is segmented into $m$ sentences $S = \{S_1, S_2, ..., S_m\}$ using the separator indicators such as period, exclamation mark and question mark. As using the period for determining the end of the sentence is ambiguous, a decision-tree-based sentence splitting method [34] is used for this purpose. As shown in Figure 3, one of the following states may occur in this decision tree.
- If there are some blank characters after a period, this period indicates the end of the sentence.
- If the period indicates an abbreviation such as "ک.م.م", the period does not show the end of the sentence.
- If the final punctuation is an exclamation or question mark, the end of the sentence is determined.

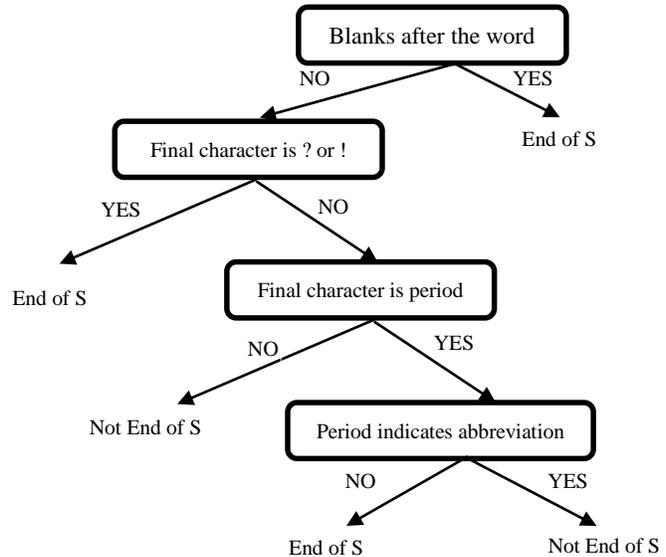

Figure 3: Decision tree for sentence segmentation in Persian, adapted from [34]; $S$ stands for sentence.

### 3-1-3- Tokenization

Tokenization is used to split a sentence into words, phrases, symbols or other meaningful tokens. Each sentence $S_i \in S$ is tokenized using the Hazm tokenizer tool [24].

### 3-1-4- Partial normalization

Partial normalization transforms plural forms to singular and removes stop words, and postfixes from words. Prefixes are not removed as they usually indicate the polarity of Persian words.

Both unigrams and bigrams are used in this step in order to extract the sentiment polarity from sentences and documents. In bigrams, both words are cleaned but only the second one is normalized. In order to accomplish this phase, we used Hazm morphology analyzer. For example, given the following sentence

میهمانان از این هتل خییییییلی لذت می برند.
Guests enjoy this hotel verrrrrrrry much".

the result would be

میهمان هتل خیلی لذت بردن
Guest hotel very enjoy

. Note that as mentioned earlier, in the bigram "لذت می برند/ enjoy", only the second word is normalized. If the word with repeated letters is polar, we increase or decrease its polarity score by a constant (e.g., +0.2). If the words with repeated letters are not polar but they are intensifiers, we double the polarity of its proceeding word, only if the proceeding word is polar.

3-2- Feature Extraction

This phase takes as input the pre-processed text and extracts features, which are required for the grouping phase. These features are listed in Table 1, which can be classified into two groups: sentence-level and document-level. Note that all features except $F_{15}$ and $F_{16}$ are used in both sentence and document-level sentiment analysis. Features $F_{15}$ and $F_{16}$ are used only in document-level. As already mentioned, these features are used in classifier combination approach, but deep learning approach implicitly extracts features by itself.

Table 1. List of features extracted from reviews, S: Sentence, D: document.

| Feature id | Name | Level |
|---|---|---|
| F1 | Avg. positive polarity of words in SentiFars | S, D |
| F2 | Avg. negative polarity of words in SentiFars | S, D |
| F3 | Avg. polarity of Positive words in Persent | S, D |
| F4 | Avg. polarity of negative words in Persent | S, D |
| F5 | Avg. polarity of positive words in LexiPers | S, D |
| F6 | Avg. polarity of negative words in LexiPers | S, D |
| F7 | Accumulative Prob. of Pos. and Neg. words | S, D |
| F8 | Accumulative Prob. of Pos. and Neg. words | S, D |
| F9 | Existence of exclamation mark | S, D |
| F10 | Existence of question mark | S, D |
| F11 | Existence of positive emoticons | S, D |
| F12 | Existence of negative emoticons | S, D |
| F13 | Domain-dependent positive indicative keywords | S, D |
| F14 | Domain-dependent negative indicative keywords | S, D |
| F15 | Polarity of the first sentence | D |
| F16 | Polarity of the last sentence | D |
| F17 | Length of document/sentence | S, D |

***Features $F_1$, $F_2$***: These features compute the average positive/negative scores of all unigrams and bigrams based on SentiFars. In SentiFars, three polarity scores summing up to one are assigned to each entry as its positivity, negativity, and objectivity. SentiFars includes over 2600 entries and polarity distribution among entries in this resource is [positive, objective, negative] = [724, 819, 1153]. Note that all words and phrases in this resource have both negative and positive scores. To the best of our knowledge, no previous work has used SentiFars for sentiment analysis in Persian.

***Features $F_3$, $F_4$:*** These features compute the average polarity of positive and negative words based on PerSent- a Persian polarity lexicon. Not similar to SentiFars, only one polarity score is assigned to each entry in PerSent. We assumed an entry in this resource as positive, if its score is greater than zero, or negative, if it is lower than zero, and neutral, otherwise. Each entry also has a POS tag. POS tagging for choosing the most relevant entry (to the context) is accomplished by Hazm Parser. PerSent includes about 1500 entries and the distribution of polar words and phrases in this resource is (positive, objective, negative) = (203,986,202).

***Features $F_5$, $F_6$***: These features are used to compute the number of positive and negative words based on LexiPers, a Persian polarity lexicon. In LexiPers, each Persian entry has a polarity tag; no polarity score is assigned to the entries. This resource includes over 6500 entries and the distribution of polar words and phrases in it is (positive, objective, negative) = (995, 4573, 1335). The characteristics of the polarity lexicons are summarized in Table 2.

Table 2. Comparison of polarity lexicons

| Corpus | Granularity level | Size (P, O, N) | Scoring form |
|---|---|---|---|
| SentiFars | Word/ Phrase | (724, 819, 1153) | (P, O, N) scores |
| PerSent | Word/ Phrase | (203, 986, 202) | Overall score |
| LexiPers | Word/ Phrase | (995, 4573, 1335) | Polarity label |

***Features $F_7$, $F_8$***: To compute these features, we defined the following equation.

$$Prob(i) = i \times P(i) \qquad (1)$$

where *i* is the number of positive/negative words in a review and *P(i)* is the probability of seeing *i* positive/negative words in positive/ negative reviews. In this equation, a review might be a document or a sentence. As we would like to measure the probability of seeing positive words in positive reviews or negative words in negative reviews, we ignored positive (or negative) reviews in computing *P(i)* for negative (or positive) class.

Figure 4 shows the probability of seeing *i* positive/ negative words in positive/negative documents. The *x* and *y* axes in this figure respectively stand for the number of positive/negative words and the probability of seeing positive/negative words in a document. As seen in Figure 4, the probability of seeing positive words in positive documents is greater than the probability of seeing negative words in negative documents. This is due to the assumption that people usually express their positive ideas more clearly than their negative ideas. Note that this diagram illustrates the computed probabilities for positive words in document and sentence levels. The same diagram could be drawn for negative words. In sentence level, the equivalent probability values are always smaller than those in the document-level. For example, the probability of seeing two positive words in a positive document (0.57) is greater than seeing two positive words in a positive sentence (0.42). Table 3 shows the computed probability values for $P(i)$s for positive and negative words in both sentence and document levels. We computed these probabilities based on manually labelled 500 Persian documents, including 3434 sentences in movie reviews, which are separated from our training and test sets.

Table 3. Probability of seeing *i* positive/negative words in positive/negative sentences or documents

| | P(i) for positive words | | P(i) for negative words | |
|---|---|---|---|---|
| i | document | sentence | document | sentence |
| 1 | 0.98 | 0.89 | 0.89 | 0.8 |
| 2 | 0.57 | 0.42 | 0.52 | 0.29 |
| 3 | 0.38 | 0.17 | 0.25 | 0.17 |
| 4 | 0.29 | 0.09 | 0.18 | 0.08 |
| 5 | 0.20 | 0.01 | 0.10 | 0.03 |
| 6 | 0.03 | 0 | 0.05 | 0 |

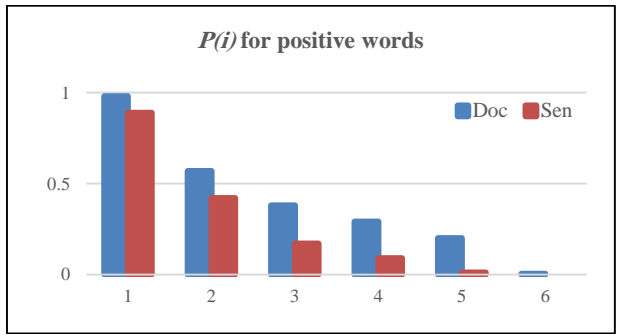

Figure 4. Probability of seeing positive words in positive documents and sentences in Persian movie reviews.

***Features $F_9$, $F_{10}$***: These features check the existence of exclamation and question marks in reviews. Generally, reviews including exclamation and question marks are more likely to be subjective. These punctuation marks are usually used to express people's surprise emotions.

***Features $F_{11}$, $F_{12}$***: These features check the existence of positive and negative emotions. Emoticons carry positive or negative emotions; therefore, they are good polarity indicators in reviews.

***Features $F_{13}$, $F_{14}$***: These features focus on domain-dependent keywords. These keywords are those words or phrases which can explicitly express the polarity of a review, in which they appear. For example, the word "کوچک (small)" usually carries negative polarity e.g., for room size in hotel domain, while it has positive polarity for battery size in camera domain. A subset of such keywords and key-phrases are listed in Table 4.

Table 4. A subset of positive and negative domain-dependent indicative keywords and key-phrases

| Positive | Negative |
|---|---|
| نزدیک /**near** | دور /far |
| بزرگ /**big** | کوچک /small |
| آرام /**calm** | شلوغ /busy |
| کنار دریا /**beside the beach** | گران /expensive |

***Features $F_{15}$, $F_{16}$***: These features are used only in document-level sentiment analysis. In these features, the average polarity of all words in the first and last sentences of a document is computed based on the scores of SentiFars and PerSent. The reason for using these features is that people usually express their ideas more explicitly in the first and last sentences of a review.

***Feature 17***: The length of a review is used as a normaliser for other features. The length of a review is computed by counting the number of its tokens.

### 3-3- Feature integration

At this phase, we prune the features in order to prepare them for the classification phase. Feature integration focuses on two challenges: handling language issues and also different granularity levels.

### 3-3-1- Handling language issues

The polarity scores of words and consequently reviews are used as features for classification; however, polarity shifters such as negation or intensification marks can modify these polarity scores. Although there exist other language issues such as covering rhetorical and sarcastic sentences, we cover only two issues: negation and intensification and leave other issues for future work. In order to handle negation, we used the following approach [5]: if the verb is negated in a sentence, all words in that sentence will be negated but if an adjective or noun is negated, only the polarity of that word will be negated. When a polar word is negated, its polarity score is shifted, meaning that this score is decreased or increased by a constant value.

On the other hand, if an adjective is intensified, only the polarity of this adjective is modified. A subset of negation marks and intensifiers are listed in Table 5. Negation and intensification handling depend on the format of polarity scores or tags assigned to each polar word/phrase. The assigned polarity might be a tag (as in LexiPers) or a float number (as in SentiFars). Algorithms 1 and 2 are proposed for negation and intensification handling when using polarity lexicons including polarity scores (SentiFars and PerSent). In the case of using polarity tags (LexiPers), the polarity of words preceding the negation mark is switched from positive to negative or from negative to neutral. The best values for const1 to const4 in these algorithms have been computed by testing different values, which resulted in 0.3, 0.2, 0.25 and 0.15 for const1 to const4, respectively.

Table 5. A subset of negators and intensifiers

| Negator | نیست، نمی باشد، نبود، نباشد، نمی شود، نخواهد بود (was/ is/ will not) |
| --- | --- |
| | بدون، بی /without |
| Intensifier | خیلی، بسیار، زیاد (additive) (very) |
| | کمی، یکم، یک ذره (reducer) (a little) |

Algorithm 1. Negation handling

```
if (using SFN)
      1.1  if (negated word is verb)
              for wi in the sentence with negated verb
                 if (wi is positive)
Pos_score(wi) - = const1;
      Neg_score(wi) += const1;
            else if (wi is negative)
Pos_score(wi) + = const1;
      Neg_score(wi) - = const1;
1.2  if (adj is negated)
            if (adj is positive)
Pos_score(adj) - = const1;
      Neg_score(adj) + = const1;
            else if (adj is negative)
Pos_score(adj) + = const1;
      Neg_score(adj) - = const1;
if (using PerSent)
      2.1  if(negated word is verb)
              for wi in the sentence with negated verb
                 if (score(wi) > 0)
            score(wi) - = const2;
            else if (score(wi) < 0)
                 score(wi) + = const2;
      2.2  if (negated word is adjective)
            if (adj is positive)
                 score(adj) - = const2;
            else if (adj is negative)
                 score(adj) + = const2;
```

Sentiment analysis can be accomplished in different granularity levels. Since we use various types of features, our approach covers word-level, phrase-level, aspect-level, sentence-level, and document-level sentiment analysis.

**Word-level sentiment analysis**

[1]http://www.sobhe.ir/hazm/

In word-level, we utilized three polarity lexicons namely SentiFars, LexiPers, and PerSent. We first cleaned and

Algorithm 2. Intensification handling

```
1. if(using SentiFars)
      1.1  if(intensifier is additive)
              if (intensified adj is positive)
Pos_score(wi) + = const3;
      Neg_score(wi) - = const3;
            else if (intensified adj is negative)
Pos_score(wi) - = const3;
      Neg_score(wi) + = const3;
      1.2  else if(intensifier is reducer)
              if (intensified adj is positive)
Pos_score(wi) - = const3;
      Neg_score(wi) + = const3;
            else if (intensified adj is negative)
Pos_score(wi) + = const3;
      Neg_score(wi) - = const3;
2. if(using PerSent)
      2.1  if(intensifier is additive)
              if (intensified adj is positive)
            score(adj) += const4;
            else if (intensified adj is negative)
            score(wi) -= const4;
      2.2  else if (intensifier is reducer)
              if (intensified adj is positive)
            score(adj) -= const4;
            else if (adj is negative)
            score(wi) - = const4;
```

3-3-2- Handling different granularities

partially normalized unigrams and bigrams and then searched them in polarity lexicons. If they were found in those lexicons, we extracted the polarity score and used it as a feature for sentiment classification.

**Phrase-level sentiment analysis**

To fulfil the phrase-level sentiment analysis, we generate phrases using a syntactic and semantic analysis module. The syntactic analysis phase uses the Hazm dependency parser[1], and generates a dependency syntactic graph $G_d$ for each sentence of the text. Figure 5 presents the dependency graph for a sample sentence. In dependency graph $G_d$, every single word is represented as a node and word-word dependencies are represented as directed edges between nodes. In other words, dependency graph $G_d$ represents binary relations between words of a sentence, in which words are connected with their parent words with a unique edge labelled with a syntactic function [2]. The sample sentence in Figure 5 is given below.

هتل استقلال تهران منظره بسیار خوبی دارد.

Tehran Esteghlal Hotel has a great view.

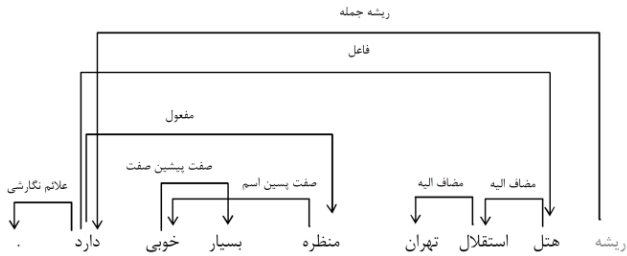

Figure 5. The dependency graph for a sample sentence

The semantic nanalysis phase takes as input the dependency graph generated by the syntactic analysis phase and augments the dependency graph with semantic information. The semantic analysis phase provides a sense mapping from surface words and named entity mentions in a dependency graph to the unique entries of ontology. In this phase, we first disambiguate the word senses using Babelfy [35]. We then filter the resulting senses by pruning the senses corresponding short tail mentions that are covered by other long-tail mentions. We map surface textual words and mentions to word senses and named entities in BabelNet ontology [35]. Figure 6 shows the semantic analysis result for a sample sentence. In Figure 6, notation *bn:in* refers to the *i*-th BabelNet sense for the given word.

| هتل استقلال تهران | منظره | بسیار خوبی | دارد |
|---|---|---|---|
| bn:17230067n | bn:00625423n | bn:00114215r | – |

Figure 6. Semantic analysis for a sample sentence

To map dependency graph's nodes to ontology entries, and create a syntactic-semantic graph, we start from the dependency graph $G_d$ of sentence s, and a set of disambiguated senses for s. If a disambiguated sense is a single token and covers a single node in $G_d$, it simply is assigned to the corresponding dependency node. If a disambiguated sense is a multi-word expression and covers more than one node in $G_d$, we merge the sub-graph referring to the same concept or entity to a single semantic node. Figure 7 shows the sense mapping and the result of the dependency graph summarization for the graph $G_d$ given in Figure 5.

The erroneous syntactic analysis of a sentence degrades the performance of later components in the syntactic analysis phase. However, we alleviate this problem by enriching the syntactic dependencies with semantic information generated by the semantic analysis phase. In order to compute the polarity of a phrase, the average polarity of all words appearing in the phrase, extracted from the dependency tree, is computed by Equation (2).

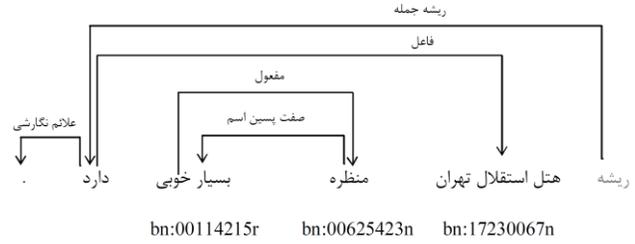

Figure 7. Syntactic-semantic analysis for a sample sentence

$$Pol(phr) = \frac{\sum_{w_i \in phr} pol(w_i)}{n} \quad (2)$$

This equation computes the average polarity of words ($w_i$s) appearing in a phrase (*phr*).

**Aspect-level sentiment analysis**

In the aspect-level phase, we used Eq. (3) to estimate the polarity of an aspect that appeared in a sentence.

$$Pol(A) = \frac{\sum pol(w_i)}{n} \quad (3)$$

*P(A)* is the average polarity of the aspect word *A*; *n* is the number of words following the aspect *A* and $w_i$ are those words proceeding *A* in the same sentence. In other words, the polarity of an aspect in a sentence is equal to the average polarity of words proceeding that aspect in the sentence. The intuition behind this method is that the polarity of an aspect is usually expressed by adjective(s) and a verb appearing after the aspect in the same sentence.

As an example, in the following sentence, the polarity of aspect word "غذا" (food) is expressed by the adjective "خوشمزه"(delicious) and the verb "دوست دارم" (like) following the aspect.

من در این هتل فقط غذای خوشمزه آن را دوست دارم.
I like only the delicious food of this hotel.

Table 6. A subset of aspect keywords in hotel domain

| (غذا)Food, (قیمت) Price,(منظره) View,(سرویس، خدمات) Service, (مکان)Place, (حمل و نقل)Transportation, (نور، روشنایی)Light, (فضا)Space … |
|---|

Subsets of aspects, which have been manually extracted from reviews in the hotel domain, are presented in Table 6. SentiFars is used to extract positive and negative scores of words in aspect-level sentiment analysis. For this aim, positive and negative polarity scores have been separately computed for each aspect, as these polarity scores have been separately assigned to each entry in SentiFars.

**Sentence and document level sentiment analysis**
In sentence-level and document-level sentiment analysis, almost the same set of features are used. Features 15 and 16 (in Table 1) are used only for the document-level. We approach a document as a bag of sentences and each sentence as a bag of words, which have been extracted from the sentences of that document. Note that first and last sentences are processed differently compared to sentences in the middle of document.

### 3-4-    Classification

Two ternary classification tasks for classifying the sentences and documents in Persian hotel reviews have been separately accomplished. Three classifiers namely, multilayer perceptron, Logistic classifier, and SMO are separately trained on 60% of data and tested on the remaining 40%. After this phase, the confidence values of these three classifiers together are used as features for training another classifier. In other words, by using the confidence values of each classifier for each class, we trained another Logistic classifier with nine feature values (3*3=9). The logistic classifier is chosen due to its higher generalization accuracy. The tool used for classification is WEKA.

### 3-5-    Deep Learning approach to Sentiment Analysis

Deep learning is a type of neural networks, which composes a more complicated and deep form of these networks. One of the characteristics of deep networks is that they can automatically extract features from data. For example, in natural language processing, deep learning methods do not require features extracted from text, like those listed in Table 1. In contrast, they automatically extract features from text and train the network; however, a big amount of data is necessary for this purpose. Word Embedding is a method which represents text by numerical vectors which can be fed to a deep neural network.

### 3-5-1-   Word Embedding

The idea of word embedding models was first proposed by Bengio [36] in 2003. A word embedding model can be simply defined as a feed-forward neural network which receives a corpus as input and provides vectors of same length (e.g., 50, 100, 200, or 300) for each word in the corpus, as output. This idea was extended in 2013, and consequently in 2014 which respectively resulted in Word2Vec [37] and Glove [38] models. FastText [39] is an extension of Word2Vect model which is used in this paper for sentiment classification of Persian reviews. Words with similar vectors would be semantically similar to each other. In other words, the smaller angle between the vectors of a word pair, the higher semantic similarity between those words. For example, the angle between "ببر/tiger" and"شیر/lion" would be smaller than the angle between "پرنده/bird" and "انسان/human".

Before using word embedding models, text is transformed to numeric format, i.e., words are replaced by numbers based on their occurrence in the corpus. Then, embedding vectors of words are extracted using a word embedding model, to train a deep learning system. This system is a polarity estimator in this work, the input of which is a list of vectors as a review (sentence or document) as well as its label in training phase, and the output is the estimated polarity of unseen reviews in the test set.

Word embedding vectors can be learned from a large corpus using one on the above-mentioned models. As this task is expensive and time-consuming, pre-trained word vectors can be also used. Word embedding vectors for different languages have been already generated for public usage. We benefit from pre-trained word vectors for Persian, generated by FastText model, available in [https://dl.fbaipublicfiles.com/fasttext/vectors-crawl/cc.fa.300.vec.gz]. These vectors have been trained using Common Crawl and Wikipedia. The training model for these vectors is Continuous Bag of Word (CBOW) in dimension 300 with character 5-grams, a window size of 5 and 10 negatives. Another model for generating word embedding vectors is Skip-gram Model. The former (CBOW) attempts to predict a word, given the context words, but the latter (Skip-gram) predicts the context words given a specific word. These models benefit from the probability of seeing a word in corpus given other words. For example in Skip-gram model[40], equation 4 is used to calculate word occurrence probabilities of context words $w_j$s, given specific word $w_t$, with window size of c. Window size indicates a window, the words inside which, are called context words. *V* in this equation indicates the vocabulary in the training corpus. More detailed explanation of these models is out of the scope of this paper.

$$\frac{1}{|V|}\sum_{t=1}^{|V|}\sum_{j=t-c, j<>t}^{t+c} \log(p(w_j|w_t)) \qquad (4)$$

Word Embedding can be used for mathematical calculation between language concepts. For example, by using word embedding vectors, equation 5 would result in the word"ملکه/queen".

زن (woman) + مرد (man) – شاه (king) =  ?          (5)

Figure 8 illustrates the 2-D form of word vectors for these concepts. Each 300-Dimensional vector of those words has been transformed to 2-Dimensional using Principal Component Analysis (PCA). As seen in this diagram, the distance and position of the concept "king" from "man" is almost the same as distance between "woman" and "queen".

### 3-5-2- Experimental setup for Deep Learning

Before using a deep learning approach, documents and sentences are pre-processed similar to the process explained in Section 3.Feature extraction in this approach is implicitly accomplished by the designed neural networks. We apply this approach only on the sentence and document levels. We used tf.keras [41] for designing and implementing a deep neural network. Keras is a deep learning API in Python. In 2019, Google has integrated the new version of TensorFnlow (tf) [42]with Keras, which is referred to as tf.keras. We used sequential model in Keras which includes five layers [43]: Define the network, Compile network, Fit network, Evaluate network, and Make predictions. The parameters used for these layers are shown in Table 6.

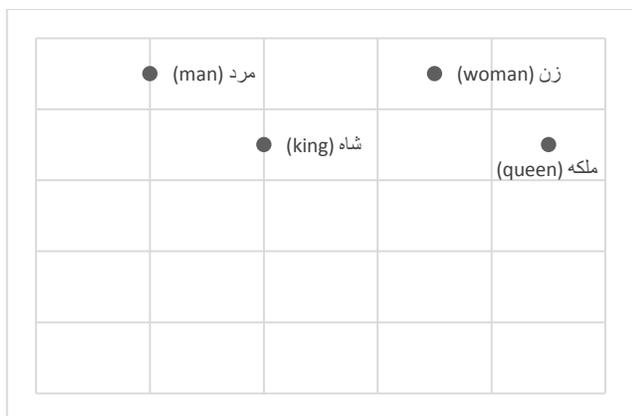

Figure 8.The 2-D form of word vectors for four Persian words

In this table, input dimensionality is the number of inputs in the visible layer of a multilayer perceptron. Hidden and output layers are the same as those in convolutional neural networks. Optimization algorithm is usually set to Stochastic gradient descent, in order to choose best value for learning rate. Finally, loss function is a quantity, which is supposed to be minimized during the training phase.

## 4- Evaluation

In this section, dataset, performance measures, obtained results, and discussion on results are presented.

Table 6. Parameters used for designing a sequential model in Keras.

| Parameter name | Value |
| --- | --- |
| Model type | Sequential |
| Input dimensionality | 2 |
| Hidden layers | 5 |
| Output layer | 1 |
| Optimization algorithm | Stochastic gradient descent (sgd) |
| Loss function | Mean squared error |

### 4-1- Dataset

The dataset used in this work is collected from Persian hotel reviews. Iranian internet users write their ideas and comments regarding hotels on websites such as booking.ir. We used 761 documents and 3613 sentences included in these documents, as Persian hotel reviews, a subset of which is available in first author's homepage[1]. These documents and sentences have been manually labelled as positive, negative, or neutral by three (plus one) native speakers. The agreement rates among three labellers in labelling the documents and sentences are respectively 81% and 84%, which means that at least two labellers agreed on the label of 81% of documents and 84% of sentences. The fourth labeller helps in labelling those reviews, which were not agreed by at least two labellers. The distribution of different classes in sentences and documents are respectively (neg, obj, pos) = (27%, 13%, 60%) and (neg, obj, pos) = (22%, 7%, 71%). The number of objective (neutral) reviews is lower than subjective ones probably because only those people write their ideas about their stay in hotels that would like to express a subjective idea.

### 4-2- Performance measures

The proposed method is evaluated using four metrics: precision (P), recall (R), accuracy (A) and $F_1$ measure [10]. Precision (P) is the number of correctly classified documents over the total number of classified documents with respect to a class. Recall (R) is the number of correctly classified documents over the total number of items that belong to a given class. Accuracy is the portion of number of correctly classified data over the number of all data.F1-

---

[1]These data are collected from booking.ir which is available in http://myweb.sabanciuniv.edu/rdehkharghani/files/2018/11/Labelled-Persian-Reviews.txt

measure is a composite score of precision and recall, computed as follows.

$$F_1 = 2 \times \left( \frac{P \times R}{P + R} \right) \quad (4)$$

## 4-3- Results

Table 7 shows the accuracy of the proposed methodology for classifier combination approach when using different subsets of features for ternary classification. The best pair of features in isolation is polarity scores extracted from SentiFars in sentence-level and polarity scores extracted from PerSent at document level. The combination of SentiFars and PerSent gives higher accuracy and the set of all features achieves the highest accuracy. The performance of features $F_{15}$ and $F_{16}$ is shown only at document-level. Moreover, as emoticons, exclamation and question marks are rarely used in reviews, features $F_9$ to $F_{12}$ are not separately evaluated. As deep learning approach does not use these features, such table cannot be provided for this approach.

Table 7. Ternary classification accuracy in the sentence- and document-level sentiment analysis on test data (40%)

| Document-level | | Sentence-level | |
|---|---|---|---|
| Feature subset | Accuracy (%) | Feature subset | Accuracy (%) |
| $F_1$-$F_2$ | 80.28 | $F_1$-$F_2$ | 69.24 |
| $F_3$-$F_4$ | 80.48 | $F_3$-$F_4$ | 68.47 |
| $F_1$-$F_4$ | 81.86 | $F_1$-$F_4$ | 69.52 |
| $F_5$-$F_6$ | 77.66 | $F_5$-$F_6$ | 67.45 |
| $F_1$-$F_6$ | 80.48 | $F_1$-$F_6$ | 69.31 |
| $F_7$-$F_8$ | 76.43 | $F_7$-$F_8$ | 67.1 |
| $F_1$-$F_8$ | 80.77 | $F_1$-$F_8$ | 70.11 |
| $F_9$-$F_{14}$ | 74.66 | $F_9$-$F_{14}$ | 65.21 |
| $F_{15}$-$F_{16}$ | 77.56 | All features | 70.81 |
| All features | 80.98 | | |

Table 8 reports precision and recall values in each class. The obtained results in all three metrics for deep learning approach are higher than the classifier combination approach, therefore, we provide the value of these metrics only for the former approach. The highest precision and recall belong to positive class probably because of the fact that people usually express their ideas more clearly when they feel positive, compared to the case they feel negative or neutral.

Figure 8 illustrates the performance of classifiers in terms of precision and recall in ternary classification tasks. The classifier combination improves the achieved precision and recall. This improvement is expected because the combination of classifiers compensates for the mistakes of one classifier by the others.

Table 8. Precision and Recall values for each class in ternary classification evaluated by test data, obtained by deep neural networks

| | Class | Precision (%) | Recall (%) | $F_1$ (%) |
|---|---|---|---|---|
| Doc-level | Negative | 82.99 | 73.05 | 77.70 |
| | Objective | 37.73 | 40.0 | 38.83 |
| | Positive | 94.44 | 92.22 | 93.31 |
| Sent-level | Negative | 89.33 | 71.52 | 79.43 |
| | Objective | 33.48 | 59.45 | 42.83 |
| | Positive | 77.67 | 89.23 | 83.04 |

In order to further clarification of classification by using the confidence values, Table 9 presents the confidence values of the classifiers for a sample sentence. In this table, Logistic classifier is more confident in assigning the negative label to the sentence; this confidence increases after combining the classifiers.

Table 9. Confidence values for a Persian sentence, obtained by different classifiers

| تنها مشکلی که داشت صبحانشون بود. | |
|---|---|
| *The only problem was the breakfast.* | |
| Classifiers | Confidence values (%) |
| Multilayer Perceptron | (Neg,Obj,Pos)= 74, 17, 9 |
| Logistic classifier | (Neg,Obj,Pos)= 77, 9, 14 |
| SMO | (Neg,Obj,Pos)= 67, 0, 33 |
| Classifier combination | (Neg,Obj,Pos)= 81, 10, 9 |

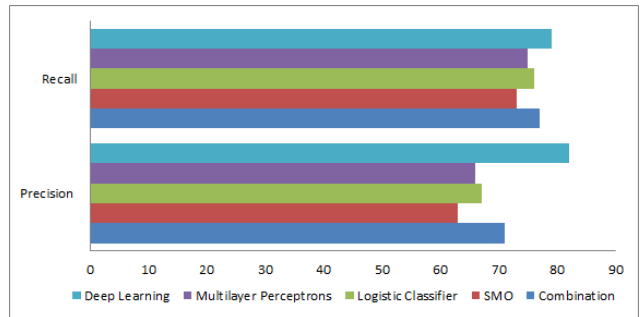

Figure 8. Precision and recall in ternary classification for different classifiers

## 4-4- Discussion and comparison

According to obtained results in Table 7 when using different sets of features, it can be concluded that using three polarity lexicons (SentiFars, PerSent, and LexiPers) achieves much higher accuracy than using features $F_9$ to $F_{14}$. As emoticons, exclamation and question marks, and domain-dependent indicative keywords are rarely used in reviews, these features achieve the lowest accuracy in isolation compared to other feature subsets. The difference between using all features and only three polarity lexicons is almost 1%. This low difference emphasizes the effect of polarity lexicons in the proposed approach.

As seen in Figure 8, Deep learning model achieves higher precision and recall than all three classifiers and also classifier combination method. Classifier combination in turn, achieves slightly higher precision and recall than the individual classifiers. This was expected as combining the predictions of several classifiers usually obtains higher performance. On the other hand, deep learning methods using word embedding have been proven to be the most effective method for text classification.

The differences of the current work with the existing ones in the literature is that the current work covers three classes (including the objective class) while the literature mostly focus on binary classification of reviews considering only the positive and negative classes. Moreover, language issues such as negation and intensification and also different granularity levels including phrase and aspect levels have not been deeply investigated in the literature.

Erroneous cases in the sentence and document-levels mostly belong to the objective class due to the lower number of objective reviews and probably because people usually do not express their neutral ideas very clearly. Another cause of the error is the lack of background knowledge in our sentiment analysis system, which is its drawback. Using common-sense and background knowledge in sentiment analysis is an open problem in most natural languages. For example, the sentence below is negative while it includes no negative word; however, people can distinguish its negativity due to their background knowledge.

بعد از یک روز اقامت در این هتل، بلافاصله هتلم را تغییر دادم.
I changed my hotel immediately after one day.

The proposed system does not have the knowledge of having negative feeling for "immediately changing the hotel after one day".

Some similar works to ours worked on Persian reviews but their domain and dataset are different from ours. Table 10 shows the results reported by these research works.

In [17], the authors accomplish a pre-processing step and then, features are extracted from the text. The extracted features are then stemmed to remove redundancy. Finally, feature selection is used to further reduce the number of features used in classification. In [19], a lexicon-based sentiment analysis framework based on the GATE pipeline [44] has been proposed. This authors utilize a statistical approach to sentiment analysis. The pipeline benefits from the existing normalization components namely, tokenizer, sentence splitter, and POS tagger. In [22], a deep learning model is proposed to extract sentiments from Persian movie reviews. The input data is passed through a pre-processing phase to perform tokenization, normalization and stemming on text. Then, the text is converted to word vectors using Fasttext library.

Table 10. The results reported in related work

| Approach | Acc. in doc. level | Acc. in sent. level | Binary/ Ternary | Dataset |
|---|---|---|---|---|
| SVM with Brown Clustering [26] | 81.08 | - | Binary | Twitter |
| Feature selection and Naïve Bayes [17] | 87.84 | | Binary | customer reviews |
| Lexicon-based S.A. pipeline [19] | 69.07 | | Binary | Movie reviews |
| Transfer Learning [45] | - | 71.87 | Ternary | Electro-devices reviews |
| RNN and LSTM [46] | 77.0 | - | Ternary | Product reviews |
| CNNs [22] | 82.86 | - | Binary | Movie reviews |
| DNN [25] | 77.95 | - | Binary | Customer reviews |

Then, for binary classification of reviews, MLP, auto-encoders and CNNs are used. In [25], a deep learning method is proposed for classifying documents based on their sentiment polarity. This method uses a word vector representation to solve most of the challenges arising from different writing styles, and the lack of the dataset in Persian by utilizing unsupervised methods. In [26], the text is first cleaned and filtered based on some keywords. Then, features are extracted from the text using the Brown Clustering algorithm. Finally, SVM classifier is used to group the reviews based on extracted features. In [45], a hybrid method is proposed, which is a combination of structural correspondence learning (SCL) and convolutional neural network (CNN). The SCL method selects the most effective pivot features, so the adaptation from one domain to similar ones cannot drop the efficiency drastically. In [46], a hybrid deep learning method for Persian sentiment analysis is proposed. In their method, the long-term dependencies are learned by long short term memory (LSTM) and local features are extracted by convolutional neural networks (CNN). They use Word2vec word representation as an unsupervised learning method.

As usually, researchers do not release their dataset, in order to provide a fair comparison, we applied some of the previous approaches on our dataset. We compared our approaches with two top-performed approaches on hotel reviews dataset. Table 11 shows the details of this comparison. The results approve the effectiveness of the proposed approach when compared with other approaches.

Table 11. Comparison of the proposed approach with similar works

| Approach | Acc. in doc. level | Acc. in sent. level | Binary/ Ternary |
|---|---|---|---|
| Deep Learning (proposed) | 84.67 | 73.06 | Ternary |
| Classifier Combination (proposed) | 80.98 | 70.81 | Ternary |
| SVM with Brown Clustering [26] | 81.08 | 70.99 | Ternary |
| CNNs [22] | 82.86 | - | Ternary |

## 5- Conclusion

In this paper, we investigated sentiment analysis in Persian and proposed a hybrid approach based on deep neural networks and classifier combination that covers some language issues such as negation and intensification and different granularity levels including word, aspect, phrase, sentence, and document-levels. The proposed approach benefits from different features and classifier combination to classify Persian reviews into opinion classes. Although the proposed approach has experimented on Persian, it can be applied to other languages with some modifications, such as updating pre-processing phase. The application of the current work includes extracting public opinion regarding an issue or product based on the reviews written by the customers, or predicting the elections by analysing the comments of internet users in social media.